\documentclass[10pt,twocolumn,letterpaper]{article}

\usepackage[accsupp]{axessibility}
 \usepackage{cvpr}              

\usepackage[dvipsnames]{xcolor}

\usepackage{multirow}
\usepackage{algorithm}
\usepackage{algorithmicx}
\usepackage{algpseudocode}
\usepackage{amsmath} 
\usepackage{url}

\definecolor{cvprblue}{rgb}{0.21,0.49,0.74}
\usepackage[pagebackref,breaklinks,colorlinks,citecolor=cvprblue]{hyperref}

\def\paperID{9921} 
\def\confName{CVPR}
\def\confYear{2024}

\title{Perturbing Attention Gives You More Bang for the Buck: Subtle Imaging Perturbations That Efficiently Fool Customized Diffusion Models}

\author{
 Jingyao Xu\textsuperscript{1}\\
 Siyang Lu\textsuperscript{1}\textsuperscript{*}\\
 \and
 Yuetong Lu\textsuperscript{1}\\
 Dongdong Wang\textsuperscript{3}\\
 \and
 Yandong Li\textsuperscript{2}\\
 Xiang Wei\textsuperscript{1}\\
 \and
 {\small \textsuperscript{1}Beijing Jiaotong University}
 {\small \textsuperscript{2}Google Research}
 {\small \textsuperscript{3}University of Central Florida}\\
 {\tt\small \{jingyaoxu, sylu\}@bjtu.edu.cn}
 }

\begin{document}

\twocolumn[{%
\renewcommand\twocolumn[1][]{#1}%
\maketitle
\begin{center}
    \centering
    \includegraphics[width=\textwidth]{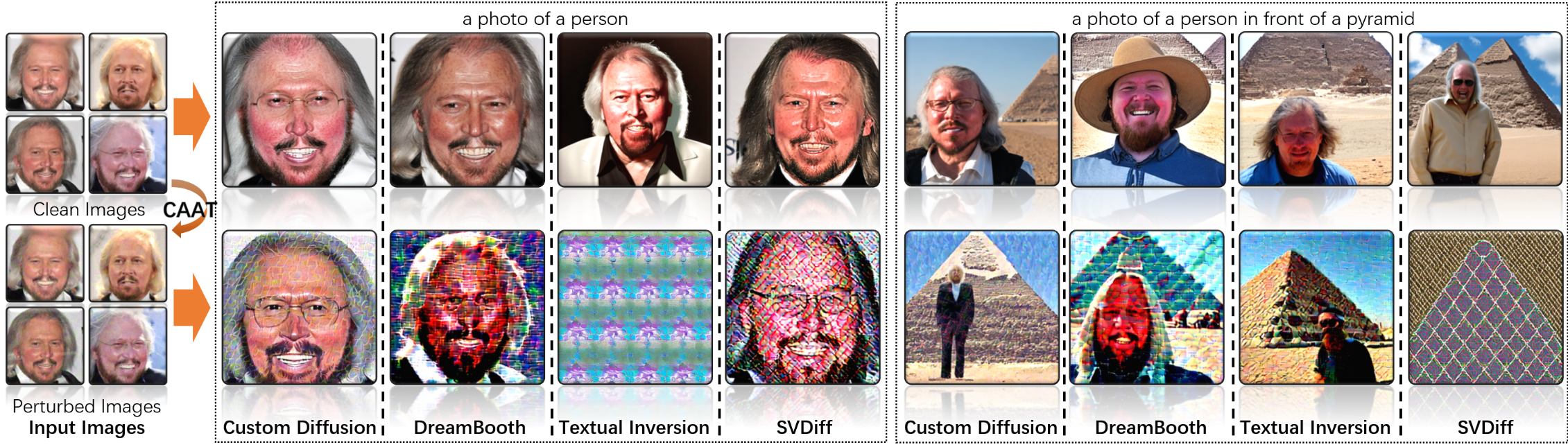}
    \captionof{figure}{With subtle perturbation, CAAT can efficiently and consistently degrade various customized diffusion models. First Line: Existing malicious attackers can use a few images publicly posted by users to generate users' images using various customized diffusion models. Second Line: Our CAAT, through subtle perturbations to the images, significantly disrupt the images generated from the customized diffusion models. We exemplify our approach by selecting the prompts ``a photo of a person" and ``a photo of a person in front of a pyramid".}
    \label{fig:CAAT}
\end{center}%
}]

\begin{abstract}
   
Diffusion models (DMs) embark a new era of generative modeling and offer more opportunities for efficient generating high-quality and realistic data samples. However, their widespread use has also brought forth new challenges in model security, which motivates the creation of more effective adversarial attackers on DMs to understand its vulnerability. We propose CAAT, a simple but generic and efficient approach that does not require costly training to effectively fool latent diffusion models (LDMs). The approach is based on the observation that cross-attention layers exhibits higher sensitivity to gradient change, allowing for leveraging subtle perturbations on published images to significantly corrupt the generated images. We show that a subtle perturbation on an image can significantly impact the cross-attention layers, thus changing the mapping between text and image during the fine-tuning of customized diffusion models. Extensive experiments demonstrate that CAAT is compatible with diverse diffusion models and outperforms baseline attack methods in a more effective (more noise) and efficient (twice as fast as Anti-DreamBooth and Mist) manner.

\end{abstract}

\section{Introduction}
\label{sec:intro}

Diffusion Models (DMs)~\cite{ho2020denoising} represent a cutting-edge advancement in the field of generative models, particularly within the realm of Text-to-Image (T2I) generation. This strategic breakthrough in generative modeling has gained recognition for its remarkable efficacy and potency in capturing intricate patterns and nuances. The technique can effortlessly transform textual descriptions into rich and visually compelling images. 



In the effort to make generative modeling with DM more efficient, the development has led to the creation of multiple variants of DM models, including Textual Inversion~\cite{gal2022image}, DreamBooth~\cite{ruiz2023dreambooth}, Custom Diffusion~\cite{kumari2023multi}, and SVDiff~\cite{han2023svdiff}, etc. These variants provide users with more customized and enhanced experiences, allowing them to obtain precise images of a specified subject by using prompts and only a small set (4-5) of relevant images. This level of customization not only empowers users to create unique and personalized content but also cultivates a widely embraced and individualized creative experience.


Despite the power of these models, users should be careful to avoid any harmful or unintended consequences that may arise from their applications. For example, malicious attackers can exploit photos available on the internet and use customized LDMs to generate deceptive and harmful fake images. Of even greater concern is the ability of attackers to fabricate false news images for their personal gain. 
Our research is dedicated to protecting users from malicious T2I attacks. Through effective strategies, we attempt to contribute valuable insights to enhance understanding and bolster security in the T2I domain. Currently, there exist adversarial attack methods, such as Anti-DreamBooth~\cite{van2023anti} and Mist~\cite{liang2023adversarial}, which have been developed to tackle the aforementioned issues. Anti-DreamBooth has exhibited remarkable proficiency in countering adversarial face attacks specifically targeted at DreamBooth, while Mist has proven its effectiveness in preserving artists' copyrights from the transformative effects of AI-for-art. However, it is crucial to acknowledge that current solutions have limitations, particularly in terms of their ability to generalize and their efficiency in terms of time.


Our research aims to overcome these limitations by focusing on the generalization of adversarial attack methods and their strong adaptability to a broader range of scenarios and systems. Moreover, the reduction of execution time is pivotal for streamlining processes, ensuring more practical and efficient applications in the real world. To tackle these challenges, we focus on attacking LDMs as a whole. A direct method involves executing a Projected Gradient Descent (PGD)~\cite{madry2017towards} attack on LDMs, targeting all parameters, similar to the approach employed by Anti-DreamBooth on DreamBooth. However, the substantial number of parameters in LDMs results in considerable time and space overhead.

\begin{figure}
  \centering
  \includegraphics[width=1\linewidth]{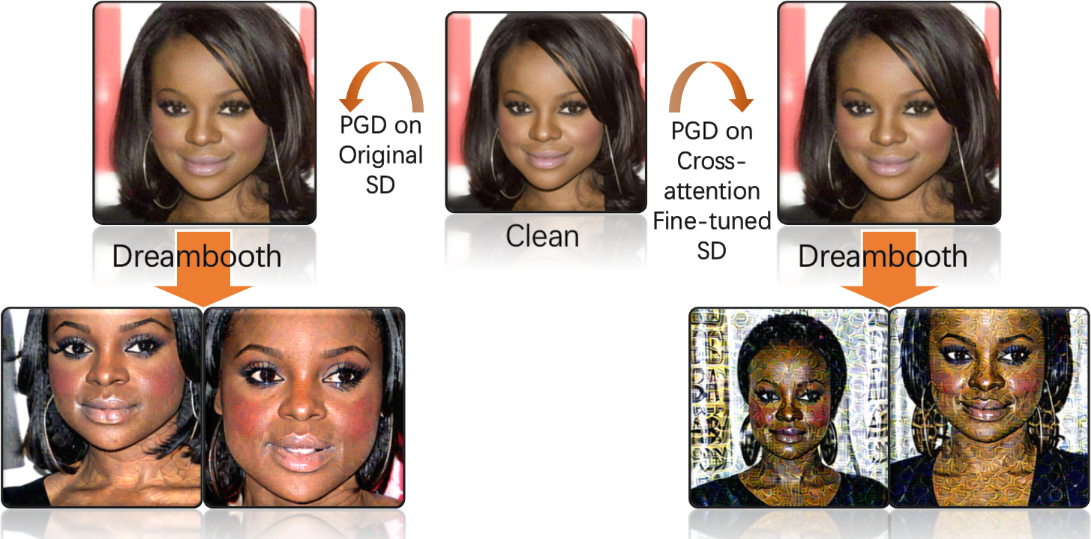}
   \caption{Illustration of the vulnerability of cross-attention layers by PGD. The comparison of adversarial attack between cross-attention and other layers reveal the vulnerability of cross-attention layers in PGD attack. We added noise to the clean images through PGD attack to generate adversarial examples. Subsequently, we employed DreamBooth~\cite{ruiz2023dreambooth} for customized fine-tuning on the adversarial examples, resulting in generated images from attacked diffusion model. }
  \label{fig:first}
\end{figure}

We propose adversarial optimization on only cross-attention layers for an efficient PGD attack. Inspired by Custom Diffusion, we have found that attention layers, specifically the cross-attention layers, play a significant role in the training process of LDMs. The cross-attention layers, integral components of LDMs, undergo substantial parameter change over the training despite having a relatively small number of parameters. We intend to leverage this observation to improve adversarial attacks on DMs. In order to investigate the effectiveness of the PGD attack on cross-attention layers of LDMs, we conduct experiments on Stable Diffusion (SD) v2.1, performing PGD attacks on both the original model and the one fine-tuned with cross-attention to produce adversarial examples. Then, we train DreamBooth on these adversarial examples to observe the attack effects. The results justify that even with minor updates to the cross-attention layers, there is a discernible improvement in the effectiveness of the attack, as illustrated in \cref{fig:first}.


According to our preliminary observations, we introduce an adversarial attack method, Cross-Attention Attack ({\bf CAAT}), that can be applied to all customized fine-tuned models based on LDMs. While adding perturbations to generate adversarial examples, we update the parameters of the cross-attention layers, a pivotal component of LDMs. The obtained perturbations will affect the cross-attention layers during fine-tuning, disrupting the mapping from text to images. Our experiments show that disrupting this key element yields significant results. \Cref{fig:CAAT} demonstrates the outstanding attack effectiveness of CAAT. Additionally, because the parameters of cross-attention layers are relatively small, our attack is lightweight and faster in training. Through extensive testing, our attack method has proven to be effective against existing LDM-based customized fine-tuning, with minimal time overhead. The overview of CAAT is presented in \cref{fig:overview}. 


In summary, our contributions are as follows:
\begin{enumerate}
    \item We identify and leverage the effectiveness of cross-attention layers in LDMs to efficient adversarial attacks on DMs.
    \item We developed CAAT, a simple yet effective attacker that exhibits excellent generalization and efficient training, providing users with robust protection against portrait rights infringements.
    \item We justify CAAT's effectiveness, efficiency, and generality through extensive experiments across various state-of-the-art adversarial attack methods on different customized LDM models. 
    \item We perform ablation study to analyze the effectiveness of influential factors for CAAT and provide suggestions for its application.
\end{enumerate}


\section{Related Work}
\label{sec:relatedwork}

\subsection{T2I models}

The advent of foundation models, as proposed by Bommasani et al.~\cite{bommasani2021opportunities}, has triggered a noticeable shift in the landscape of deep learning. This transition is characterized by a growing emphasis on large-scale models, housing billions of parameters, and trained on extensive datasets. This evolution has, in turn, propelled advancements in T2I generation models. In the domain of image generation, Generative Adversarial Networks (GANs)\cite{goodfellow2020generative}, once emblematic and canonical, are witnessing a gradual displacement by diffusion models\cite{ho2020denoising}. This shift is attributed to the disruptive influence of diffusion models on the traditional structure of GANs, leading to a notable improvement in the quality of generated content~\cite{dhariwal2021diffusion}. The remarkable success of diffusion models in the realm of image generation~\cite{balaji2022ediffi,ramesh2022hierarchical,saharia2022photorealistic,nichol2021glide} has redirected research interests, with an increasing focus on their potential applications in T2I generation. 

As an exemplar T2I model, Stable Diffusion (SD)\cite{rombach2021highresolution} adopts the Contrastive Language-Image Pre-training (CLIP)\cite{radford2021learning} Text Encoder for encoding textual information. SD generates a Gaussian noise matrix, employing a random function as a ``substitute" for the Latent Feature. This matrix is then fed into the ``image optimization module" of the SD model, featuring a U-Net~\cite{ronneberger2015u} network. The U-Net network is tasked with predicting noise while simultaneously integrating semantic information from the textual input. The Scheduler refines the noise predicted by the U-Net at each iteration. Finally, this refined information undergoes processing in the Variational AutoEncoder (VAE)~\cite{kingma2013auto} Decoder, culminating in the generation of the final image. This paradigmatic shift in image generation models underscores the dynamic nature of the field, spurred by the adoption of foundation models and the continuous pursuit of enhanced capabilities.


Dreambooth~\cite{ruiz2023dreambooth} is realized through SD fine-tuning, employing a minimal input of several images (typically three or four images) to generate corresponding images under various prompts. Diverging from the SD fine-tuning approach, Textual Inversion~\cite{gal2022image} operates with a reduced set of images, generating and training images with a similar style by specifying new keywords. SVDiff~\cite{han2023svdiff} presents a lighter-weight diffused fine-tuning model designed to mitigate the risks of overfitting and language drift simultaneously. In contrast, Custom Diffusion~\cite{kumari2023multi} optimizes solely the parameters in the cross-attention layers of the T2I diffusion model. This targeted optimization facilitates the efficient learning of new concepts, surpassing DreamBooth and Textual Inversion models in terms of image generation performance. Notably, Custom Diffusion achieves this while minimizing memory overhead and enhancing inference efficiency.

\subsection{Adversarial attack}

The emergence of the Fast Gradient Sign Method (FGSM)~\cite{goodfellow2014explaining} has led researchers in the field of machine learning to focus a significant portion of their attention on adversarial attacks. Its idea is to add the computed loss value to the input image, causing an increase in the loss value of the network's output and ultimately leading to incorrect model predictions. This has also spurred the development of other similar methods~\cite{shafahi2019adversarial,kurakin2018adversarial,madry2017towards,zhang2019you,zhu2019freelb,jiang2019smart}. BIM~\cite{kurakin2018adversarial}, through multiple iterations, introduces small perturbations along the direction of increasing gradients, resulting in more accurate perturbations compared to FGSM. Among them, Projected Gradient Descent (PGD)~\cite{madry2017towards} deserves special attention. Unlike the fast attack method of FGSM that operates with a single iteration, PGD is an iterative attack method. It generates stronger adversarial samples by performing multiple iterations, allowing it to bypass certain defense mechanisms.

\subsection{User safeguarding through image cloaking}
Image cloaking is a widely researched field due to its importance in maintaining privacy and preventing misuse of images. Pixelization and blurring are commonly used techniques for hiding personal information such as faces and license plates. With the continuous development of the T2I model, people are paying attention to its remarkable image generation capabilities while also being wary of the potential risks of its misuse. Therefore, when faced with malicious attacks on images, we should take measures to prevent their success. For T2I, our goal is to introduce imperceptible perturbations into pre-existing images before their release. Once these images are reused, the model will generate images with negative effects. \cite{ shan2023glaze, salman2023raising, zhao2023unlearnable, ma2023generative, cui2023ft} attempt to prevent images from being edited or exploited.

Similar to our objective, Anti-DreamBooth~\cite{van2023anti} and Mist~\cite{liang2023adversarial} aim to disrupt the generative modeling quality by adding subtle perturbations to images. Our work differs from their method of adding perturbations to images while keeping the parameters unchanged. In our approach, we train and update the parameters of the cross-attention layers of the model. By updating these parameters, we introduce perturbations to the image.
 
\section{Method}
\label{sec:Method}

        \begin{figure}[t]
          \centering
           \includegraphics[width=1\linewidth]{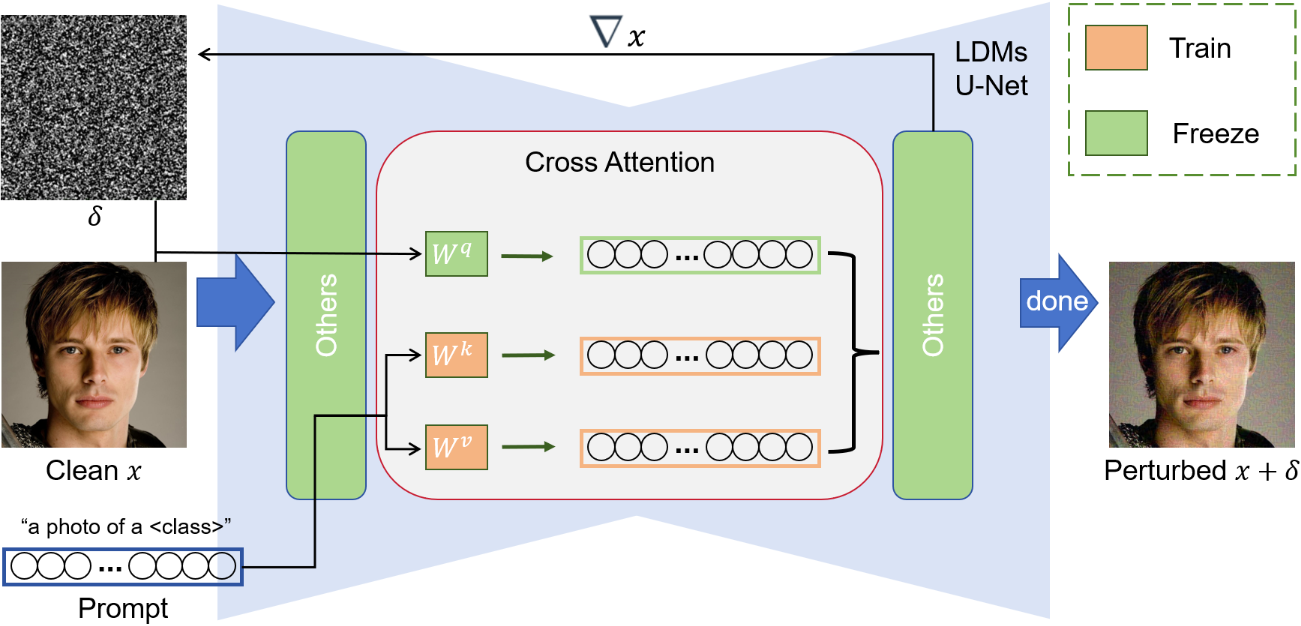}
        
           \caption{Schematic of CAAT attacking a T2I diffusion model. During attacker training, first, $W_K$ and $W_V$ of the cross-attention layer are optimized. Then, the perturbation $\delta$ is optimized based on the gradient of $x$, yielding perturbed image $x+\delta$. }
           \label{fig:overview}
        \end{figure}

\Cref{fig:overview} illustrates an overview of our CAAT approach that leverages PGD training on cross-attention layer to effectively attack LDMs. We introduce the principles of diffusion models and adversarial attacks in ~\cref{subsec:Diffusion} and  ~\cref{subsec:Adversarial}. Furthermore, CAAT is proposed in detail in \cref{subsec:CAAT}.

\subsection{Diffusion models}\label{subsec:Diffusion}
In the current field of Text-to-Image (T2I), diffusion models have established themselves as the reigning champions, capable of generating diverse and realistic images. Diffusion models include two processes: a forward process and a backward process. The forward process gradually introduces noise into the input image until the data distribution becomes pure Gaussian noise, while the backward process learns in reverse, extracting the desired data from random noise.
Given the input image $x_{0}\sim q(x)$, forward process injects noise into $x_{0}$ over $T$ steps, resulting in a Markov chain ${x_1,x_2,x_3,...,x_T}$, each $x_{t}$ satisfies:
\begin{equation}
x_{t}=\sqrt{\alpha_t}x_{0}+\sqrt{1-\alpha_t}\epsilon , 
    \label{eq:eq_DM_forward}
\end{equation} 
where $\epsilon\sim\mathcal{N}(0,\mathbf{I})$ and $\alpha_t$ is obtained by a noise scheduler.

Given noise image $x_{t}$ in time $t$, backward process learn to denoise the noise image to obtain $x_{t-1}$. The training objective of diffusion models can be expressed succinctly as follows:
\begin{equation}
\mathcal{L}_{DM}(\theta,x_{0})=\mathbb{E}_{x_{0},t,\epsilon}\|\epsilon-\epsilon_{\theta}(x_{t},t)\| ,
    \label{eq:eq_DM}
\end{equation} 
where $\epsilon_{\theta}$ is a parametric neural network.

The prompt-based diffusion models, such as LDMs, have an additional prompt $c$ to generate images that better match the text description. After undergoing encoding processing, the prompt $c$ is mapped to the intermediate layers in the U-Net of LDMs through the introduction of cross-attention layers, achieving the mapping from text to images:

        \begin{table*}[t]
        \centering
        \caption{Effectiveness assessment with four evaluation metrics by comparing different attackers on different T2I diffusion models. {\bf Bold} is the best score and \underline{underlining} is the second-best score. CAAT achieved  {\bf 12 best} and \underline{3 second-best} out of 16 metrics, demonstrating its superior attack effectiveness. The abbreviation for ``Anti-dreambooth" is denoted as ``Anti".}
          \resizebox*{\linewidth}{!}{
        \begin{tabular}{c|cccccccccccccccc}
        \hline
                                       & \multicolumn{16}{c}{T2I diffusion models}                                                                                                                          \\ \cline{2-17} 
        \multirow{2}{*}{attack} & \multicolumn{4}{c|}{Custom Diffusion}   & \multicolumn{4}{c|}{DreamBooth}         & \multicolumn{4}{c|}{SVDiff}             & \multicolumn{4}{c}{Textual Inversion} \\
                                       & FR  $\downarrow$& FS  $\downarrow$& IR  $\downarrow$& \multicolumn{1}{c|}{FID $\uparrow$} & FR  $\downarrow$& FS  $\downarrow$& IR  $\downarrow$& \multicolumn{1}{c|}{FID $\uparrow$} & FR  $\downarrow$& FS  $\downarrow$& IR  $\downarrow$& \multicolumn{1}{c|}{FID $\uparrow$} & FR  $\downarrow$& FS  $\downarrow$& IR  $\downarrow$& FID $\uparrow$\\ \hline
        clean                          &    1.00&    0.52&    0.47& \multicolumn{1}{c|}{195}    &    0.98&    0.52&    0.53& \multicolumn{1}{c|}{179}    &    0.99&    0.57&    0.68& \multicolumn{1}{c|}{218}    &         1.00&         0.47&         0.27&         242\\ \hline
        Anti                &    1.00&    0.48&    0.21& \multicolumn{1}{c|}{207}    &    \underline{0.81}&    0.40&    0.16& \multicolumn{1}{c|}{\underline{307}}    &    0.94&    0.38&    0.46& \multicolumn{1}{c|}{308}    &          \underline{0.50}&         \underline{0.15}&         \underline{-0.92}&         \underline{378}\\ \hline
        Mist                           &    1.00&    {\bf 0.39}&    \underline{0.03}& \multicolumn{1}{c|}{\underline{233}}    &    0.99&    {\bf 0.32}&    \underline{0.13}& \multicolumn{1}{c|}{275}    &    \underline{0.88}&    {\bf 0.21}&    {\bf -0.12}& \multicolumn{1}{c|}{\underline{317}}    &         0.63&         {\bf 0.14}&         -0.85&         348\\ \hline
        CAAT                      &    1.00&    \underline{0.42}&    {\bf -0.36}& \multicolumn{1}{c|}{{\bf 250}}    &    {\bf 0.64}&    {\bf 0.32}&    {\bf -0.14}& \multicolumn{1}{c|}{{\bf 371}}    &    {\bf 0.85}&    \underline{0.34}&    \underline{0.29}& \multicolumn{1}{c|}{{\bf 355}}    &         {\bf 0.43}&         {\bf 0.14}&         {\bf -1.30}&         {\bf 396}\\ \hline
        \end{tabular}%
        }
        \label{tab:main_tab}
        \end{table*}

        \begin{table*}[t]
        \centering
        \caption{Effectiveness analysis across varying noise budgets for CAAT on selected T2I diffusion models, where ``*" denotes the default budget setting.}
        \resizebox*{\linewidth}{!}{
        \begin{tabular}{c|cccccccccccccccc}
        \hline
                                       & \multicolumn{16}{c}{T2I diffusion models}                                                                                                                          \\ \cline{2-17} 
        \multirow{2}{*}{$\eta$} & \multicolumn{4}{c|}{Custom Diffusion}   & \multicolumn{4}{c|}{DreamBooth}         & \multicolumn{4}{c|}{SVDiff}             & \multicolumn{4}{c}{Textual Inversion} \\
                                       & FR  $\downarrow$& FS  $\downarrow$& IR  $\downarrow$& \multicolumn{1}{c|}{FID $\uparrow$} & FR  $\downarrow$& FS  $\downarrow$& IR  $\downarrow$& \multicolumn{1}{c|}{FID $\uparrow$} & FR  $\downarrow$& FS  $\downarrow$& IR  $\downarrow$& \multicolumn{1}{c|}{FID $\uparrow$} & FR  $\downarrow$& FS  $\downarrow$& IR  $\downarrow$& FID $\uparrow$\\ \hline
        0.05                          &    1.0&    0.45&    -0.27& \multicolumn{1}{c|}{225}    &    0.95&    0.44&    0.44& \multicolumn{1}{c|}{251}    &    0.98&    0.44&    0.51& \multicolumn{1}{c|}{307}    &         0.83&         0.34&         -0.17&         302\\ \hline
        *0.10                     &    1.0&    0.42&    {\bf -0.36}& \multicolumn{1}{c|}{250}    &     0.64&    0.32&    -0.14& \multicolumn{1}{c|}{371}    &    0.85&    0.34&    0.29& \multicolumn{1}{c|}{355}    &          0.51&         0.14&         {\bf -1.30}&         396\\ \hline
        0.15                &    {\bf 0.95}&    {\bf 0.38}&    -0.24& \multicolumn{1}{c|}{{\bf 284}}    &    {\bf 0.34}&    {\bf 0.28}&    {\bf -0.58}& \multicolumn{1}{c|}{{\bf 401}}    &    {\bf 0.76}&    {\bf 0.29}&    {\bf 0.15}& \multicolumn{1}{c|}{{\bf 390}}    &         {\bf 0.50}&         {\bf 0.09}&         -0.90&         {\bf 405}\\ \hline
        
        \end{tabular}
        }
        \label{tab:ablation_tab}
        \end{table*}

\begin{equation}
\mathcal{L}_{LDM}(\theta,x_{0})=\mathbb{E}_{x_{0},t,\epsilon,c}\|\epsilon-\epsilon_{\theta}(x_{t},t,c)\| .
    \label{eq:eq_CDM}
\end{equation} 

\subsection{Adversarial attack}\label{subsec:Adversarial}
Adversarial attack, now prevalent in various domains, represents a sophisticated and pervasive class of attack methods in the contemporary digital landscape. It were initially introduced for targeting classification models whose attacks leverage carefully crafted inputs with the aim of deceiving, misleading, or undermining classification models, thereby compromising their performance or inducing misclassifications. In short, it finds an alternative input $x^{\prime}$ for a given input $x$ and its label that causes it not to be classified as its true label which can be formulated by
\begin{equation}
\begin{aligned}x^{\prime}&:=\arg\max_{x^{\prime}}\mathcal{L}_\theta(x^{\prime}) ,\\s.t.&\quad||x-x^{\prime}||\le\eta,\end{aligned}
    \label{eq:adv}
\end{equation} 
where $\mathcal{L}_\theta$ is a classification network and $\eta$ is a small positive constant, ensuring that $x^{\prime}$ does not deviate too far from $x$.

\subsection{CAAT}\label{subsec:CAAT}
It is critical to analyze the vulnerability of DMs from the perspectives of both effectiveness and efficiency. {\bf CAAT} is proposed based on this goal and developed upon PGD attack. To prevent the misuse of customized diffusion models, we aim to obtain adversarial examples that, when used as inputs for customized LDMs models, result in the fine-tuned model losing the ability to generate images corresponding to specific themes, thereby disrupting the quality of the T2I generate images. To achieve this, we introduce a perturbation $\delta$ into the input image $x$, which is visually imperceptible controlled by $\eta$, making it impossible for the model to learn useful information during training. The overall objective of CAAT can be formulated by
\begin{equation}
\begin{aligned}\delta&:=\arg\max_\delta\mathcal{L}_{LDM}(\theta,x+\delta),\\&\quad\text{where}~\|\delta\|\leq\eta.\end{aligned}   
    \label{eq:adv_DM}
\end{equation} 

A classic and practical adversarial attack method is the PGD attack, which is applied to Anti-DreamBooth and Mist. PGD is applied to a trained model, obtaining gradients during the attack process without updating model parameters. Different from this convention, we optimize LDMs during the CAAT training. By this means, LDMs is trained on adversarial examples $x+\delta$ to enhance its robustness. Simultaneously, adversarial examples are applied to a more robust LDMs, leading to the generation of adversarial examples with improved attack effectiveness. The LDMs training process of CAAT can be formulated by
\begin{equation}
\begin{aligned}\theta&:=\arg\min_\theta\mathcal{L}_{LDM}(\theta,x+\delta),\\&\quad\text{where}~\|\delta\|\leq\eta.\end{aligned}   
    \label{eq:adv_DM}
\end{equation}

We leverage the observations and best practices in Custom Diffusion~\cite{kumari2023multi} to analyze and select the layers for efficient attack. During the fine-tuning of diffusion models, cross-attention layers have the fewest parameters but undergo the most changes. This observation indicates cross-attention layer plays a significant role in model optimization during the training process. Conventional attention between images and texts can be formulated as follows:
\begin{equation}
    \mathrm{Attention}(Q,K,V)=\mathrm{softmax}\left(\frac{QK^T}{\sqrt{d}}\right)\cdot V,
    \label{eq:coss-atten}
\end{equation}
where $Q=W_{Q}\mathbf{f},K=W_{K}\mathbf{c},V={W_{V}}\mathbf{c}$. Here, $\mathbf{f}\in\mathbb{R}^{(h\times w)\times l}$ is image features, $\mathbf{c}\in\mathbb{R}^{s\times d}$ is features of prompt and $W_{Q}$, $W_{K}$, and $W_{V}$ are learnable matrices that respectively map the input to query, key and value. $W_{Q}$ processes the image input features, while $W_{K}$ and $W_{V}$ handle the text input features. Disrupting the learning of $W_{K}$ and $W_{V}$ can undermine the mapping between text and images in customized fine-tuned diffusion models. After this undermining, the model is capable of recognizing or acknowledging the content or subject of the image, but it lacks the ability to categorize or classify it into specific groups or types. Therefore, we update the parameters $W_{K}$ and $W_{V}$ of cross-attention layers.

In summary, during the CAAT process, we freeze the model parameters other than $W_{K}$ and $W_{V}$ and only update them to facilitate the learning of the mapping between text input and image input by the model. Simultaneously, we search for a perturbation $\delta$ in images $x$ that causes the model to lose the aforementioned capabilities. (The reasons for choosing to simultaneously update parameters and add noise are discussed in detail in Supplementary Material \hyperref[sec:separated_optimization]{D}). The search process is implemented by calculating the gradients for $x$ and performing gradient ascent. The algorithm is introduced in \cref{algorithm:CAAT}.

\renewcommand{\algorithmicrequire}{\textbf{Input:}}
\renewcommand{\algorithmicensure}{\textbf{Output:}}
\renewcommand{\algorithmicrepeat}{\textbf{Repeat }}
 \begin{algorithm}
        \caption{CAAT}
        \begin{algorithmic}[1]
            \Require Images $x$, K layers parameter $W_K$, V layers parameter $W_V$, step length $\alpha$, limitation $\eta$, steps number $N$, LDMs learning rate $l$
            \Ensure Perturbed images $x^\prime$
            \State Initialize $\delta$
            \For{$i = 1 \to N$} 
                \State $\nabla_{K},\nabla_{V},\nabla_{x}\leftarrow\mathcal{L}_{LDM}((W_{K},W_{V}),x+\delta)$
                \State $W_K\leftarrow W_K-l\nabla_{K}$
                \State $W_V\leftarrow W_V-l\nabla_{V}$
                \State $\delta\leftarrow\delta+\alpha\mathrm{sgn}\nabla_{x}$\Comment{$\nabla_{x}$ is from the input images.}
                \If{$||\delta|| > \eta$}
                    \State $\delta\leftarrow clip(\delta,-\eta,\eta)$\Comment{limit $||\delta||$ within [0, $\eta$]} 
                \EndIf
            \EndFor
            \State $x^\prime\leftarrow x+\delta$
            \State \Return{$x^\prime$}
        \end{algorithmic}
        \label{algorithm:CAAT}
    \end{algorithm}

\section{Experiments}
\label{sec:experiments}

        \begin{table*}[]
        \centering
        \caption{Effectiveness evaluation across different LDMs versions on different T2I diffusion models. CAAT is trained on SD v2.1.}
        \resizebox*{\linewidth}{!}{
        \begin{tabular}{c|c|cccccccccccccccc}
        \hline
        \multirow{3}{*}{version} & \multirow{3}{*}{attack} & \multicolumn{16}{c}{T2I diffusion models}                                                                                                                          \\ \cline{3-18} 
                                 &                         & \multicolumn{4}{c|}{Custom Diffusion}   & \multicolumn{4}{c|}{DreamBooth}         & \multicolumn{4}{c|}{SVDiff}             & \multicolumn{4}{c}{Textual Inversion
        } \\
                                 &                         & FR  $\downarrow$& FS  $\downarrow$& IR  $\downarrow$& \multicolumn{1}{c|}{FID $\uparrow$} & FR  $\downarrow$& FS  $\downarrow$& IR  $\downarrow$& \multicolumn{1}{c|}{FID $\uparrow$} & FR  $\downarrow$& FS  $\downarrow$& IR  $\downarrow$& \multicolumn{1}{c|}{FID $\uparrow$} & FR  $\downarrow$& FS  $\downarrow$& IR  $\downarrow$& FID $\uparrow$\\ \hline
        \multirow{2}{*}{v1.4}    & clean                   &    0.97&    0.53&    0.21& \multicolumn{1}{c|}{224}    &    0.96&    0.47&    0.50& \multicolumn{1}{c|}{197}    &    0.93&    0.50&    0.23& \multicolumn{1}{c|}{248}    &         0.97&         0.41&         0.17&         248\\
                                 & CAAT                &    {\bf 0.79}&    {\bf 0.45}&    {\bf -0.24}& \multicolumn{1}{c|}{{\bf 313}}    &    {\bf 0.65}&    {\bf 0.27}&    {\bf -0.34}& \multicolumn{1}{c|}{{\bf 350}}    &    {\bf 0.16}&    {\bf 0.25}&    {\bf -1.57}& \multicolumn{1}{c|}{{\bf 447}}    &         {\bf 0.15}&         {\bf 0.07}&         {\bf -1.94}&         {\bf 501}\\ \hline
        \multirow{2}{*}{v1.5}    & clean                   &    0.96&    0.54&    0.27& \multicolumn{1}{c|}{217}    &    0.97&    0.46&    0.41& \multicolumn{1}{c|}{199}    &    0.97&    0.52&    0.26& \multicolumn{1}{c|}{233}    &         0.90&         0.41&         -0.07&         259\\
                                 & CAAT                &    {\bf 0.88}&    {\bf 0.46}&    {\bf -0.03}& \multicolumn{1}{c|}{{\bf 284}}    &    {\bf 0.49}&    {\bf 0.31}&    {\bf -0.50}& \multicolumn{1}{c|}{{\bf 364}}    &    {\bf 0.12}&    {\bf 0.24}&    {\bf -1.41}& \multicolumn{1}{c|}{{\bf 425}}    &         {\bf 0.31}&         {\bf 0.09}&         {\bf -1.48}&         {\bf 441}\\ \hline
        \end{tabular}
        }
        \label{tab:SD}
        \end{table*}
    
        \begin{table*}[]
        \centering
        \caption{Effectiveness assessment by varying the number of perturbed images. The results demonstrate the proportion of perturbed images obtained by CAAT that affect the image quality of the T2I diffusion models, considering four input images. }
        \resizebox*{\linewidth}{!}{
        \begin{tabular}{c|c|cccccccccccccccc}
        \hline
        \multirow{3}{*}{Clean} & \multirow{3}{*}{Perturbed} & \multicolumn{16}{c}{T2I diffusion models}                                                                                                                          \\ \cline{3-18} 
                               &                            & \multicolumn{4}{c|}{Custom Diffusion}   & \multicolumn{4}{c|}{DreamBooth}         & \multicolumn{4}{c|}{SVDiff}             & \multicolumn{4}{c}{Textual Inversion
        } \\
                               &                            & FR  $\downarrow$& FS  $\downarrow$& IR  $\downarrow$& \multicolumn{1}{c|}{FID $\uparrow$} & FR  $\downarrow$& FS  $\downarrow$& IR  $\downarrow$& \multicolumn{1}{c|}{FID $\uparrow$} & FR  $\downarrow$& FS  $\downarrow$& IR  $\downarrow$& \multicolumn{1}{c|}{FID $\uparrow$} & FR  $\downarrow$& FS  $\downarrow$& IR  $\downarrow$& FID $\uparrow$\\ \hline
        4                      & 0                          &    1.00&    0.52&    0.47& \multicolumn{1}{c|}{195}    &    0.98&    0.52&    0.53& \multicolumn{1}{c|}{179}    &    0.99&    0.57&    0.68& \multicolumn{1}{c|}{218}    &         1.00&         0.47&         0.27&         242\\ \hline
        3                      & 1                          &    1.00&    0.50&    0.16& \multicolumn{1}{c|}{202}    &    0.97&    0.50&    0.53& \multicolumn{1}{c|}{197}    &    0.98&    0.53&    0.61& \multicolumn{1}{c|}{234}    &         1.00&         0.44&         0.24&         252\\
        2                      & 2                          &    1.00&    0.47&    0.09& \multicolumn{1}{c|}{207}    &    0.91&    0.50&    0.26& \multicolumn{1}{c|}{249}    &    0.94&    0.47&    0.60& \multicolumn{1}{c|}{247}    &         0.88&         0.37&         -0.06&         281\\
        1                      & 3                          &    1.00&    0.45&    -0.15& \multicolumn{1}{c|}{228}    &    0.76&    0.44&    0.10& \multicolumn{1}{c|}{301}    &    0.93&    0.39&    0.48& \multicolumn{1}{c|}{294}    &         0.71&         0.24&         -0.44&         314\\ \hline
        0                      & 4                          &    1.00&    {\bf 0.42}&    {\bf -0.36}& \multicolumn{1}{c|}{{\bf 250}}    &    {\bf 0.64}&    {\bf 0.32}&    {\bf -0.14}& \multicolumn{1}{c|}{{\bf 371}}    &    {\bf 0.85}&    {\bf 0.34}&    {\bf 0.29}& \multicolumn{1}{c|}{{\bf 355}}    &         {\bf 0.43}&         {\bf 0.14}&         {\bf -1.30}&         {\bf 396}\\ \hline
        \end{tabular}
        }
        \label{tab:clean_perb}
        \end{table*}

        \begin{figure*}[t]
          \centering
           \includegraphics[width=1\linewidth]{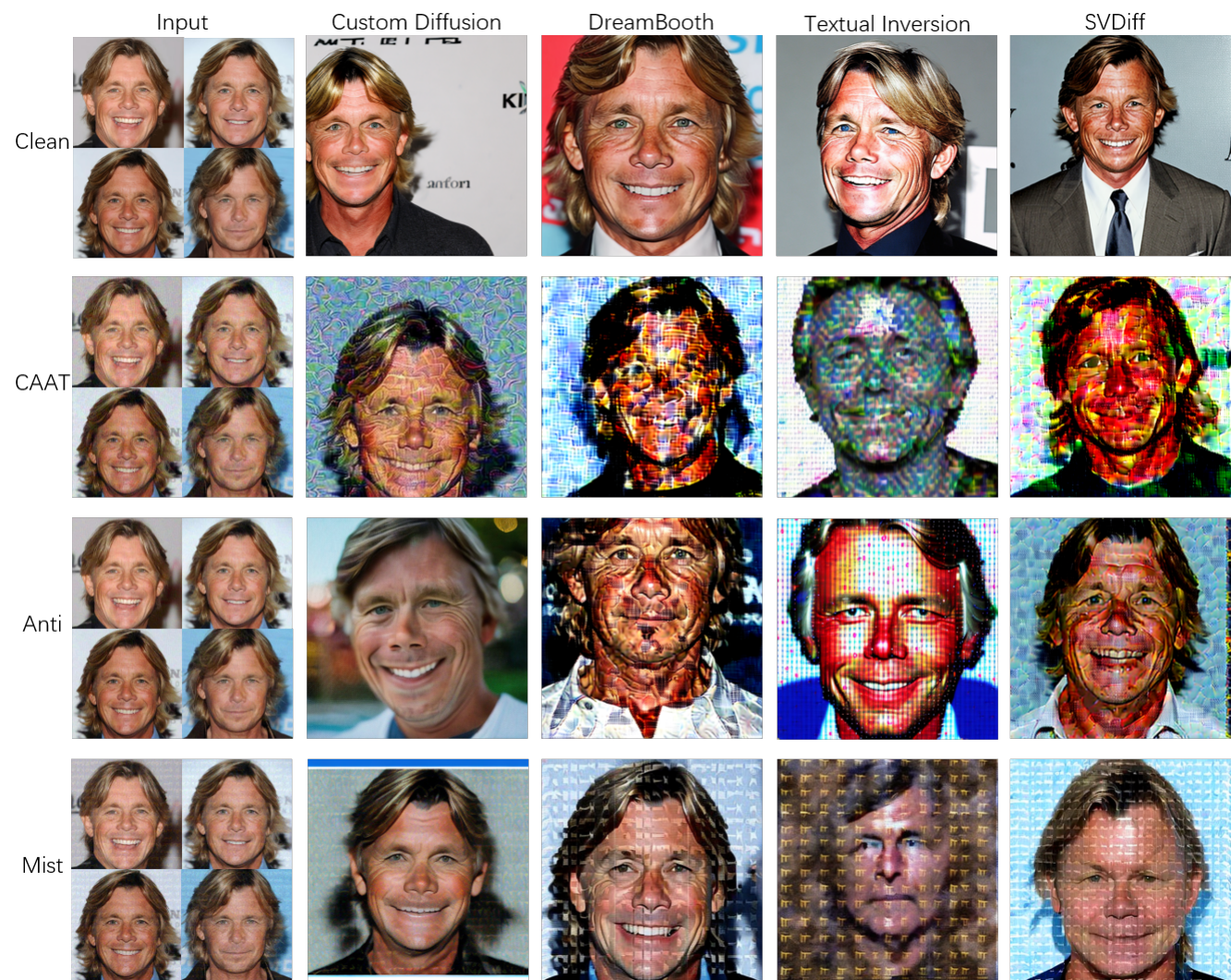}
        
           \caption{Comparison in the images generated by different T2I diffusion models with different attackers. The first column illustrates the four input images. For attackers by row, the observation of the perturbation pattern can refer to \cref{fig:perturb}. For diffusion models by column, four models are selected and compared to evaluate the performance of attackers.}
           \label{fig:compare}
        \end{figure*}

        \begin{figure}[t]
          \centering
           \includegraphics[width=0.95\linewidth]{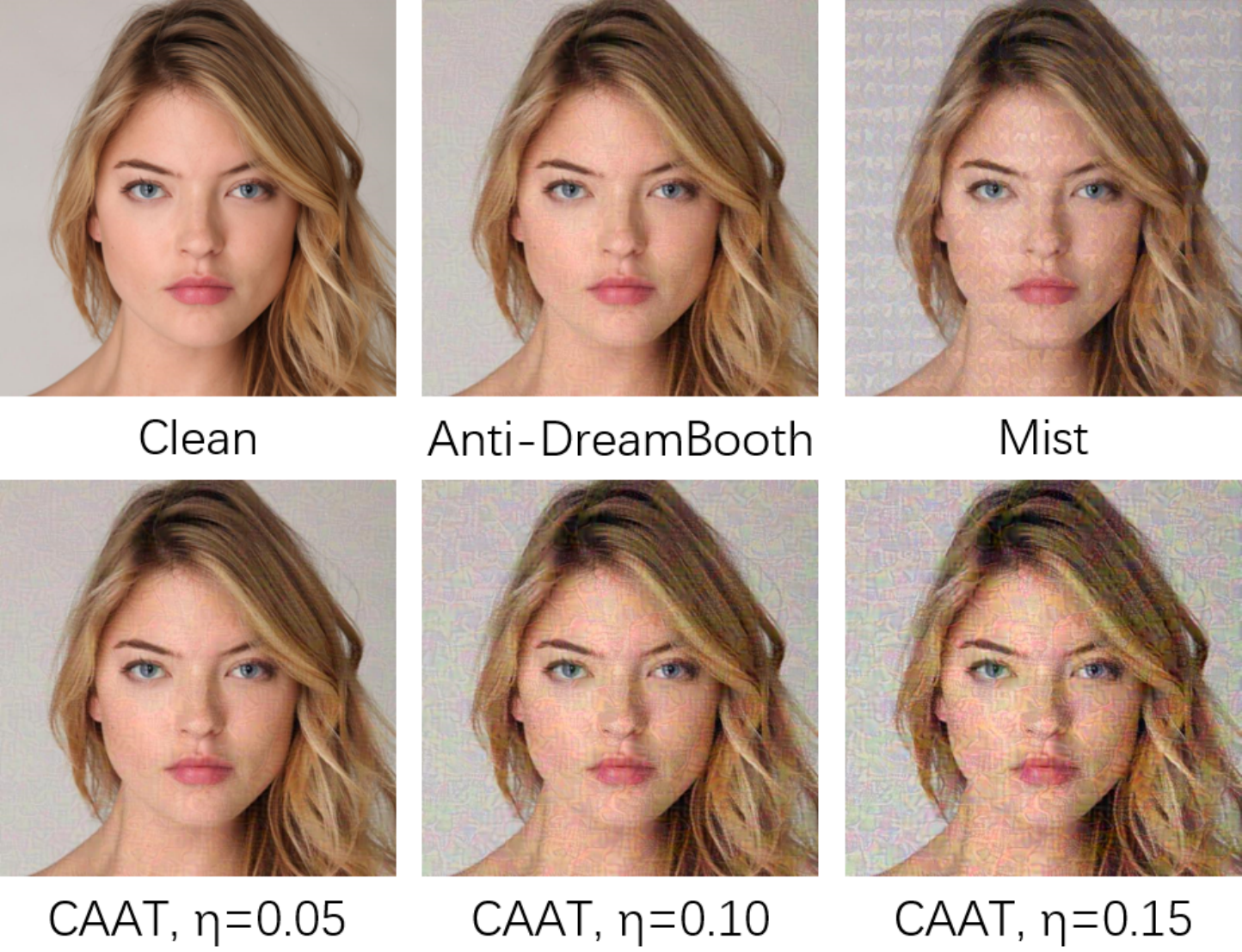}
           \caption{Adversarial examples of different attacker after adding noise. The parameter configurations of Anti-DreamBooth and Mist follow the default settings in \cref{tab:attacks}.}
           \label{fig:perturb}
        \end{figure}

In this section, we valuate the effectiveness of CAAT on customized LDMs through experiments. Specifically, we compare CAAT with other attack methods across various DMs to evaluate the effectiveness, generalization, and efficiency of CAAT. 

\subsection{Experimental setup}

{\bf Datasets.} To justify our proposed CAAT attacker, the testing datasets should comply the following criteria: 1) an ample supply of face images, 2) categorized based on individuals, and 3) high-quality images with high resolution. In accordance with these criteria and taking inspiration from Anti-DreamBooth, we select the face datasets CelebA-HQ~\cite{karras2017progressive}. CelebA-HQ is the high-resolution version of CelebA~\cite{liu2015deep}, which includes 10,177 unique celebrity identities and 202,599 face images. CelebA-HQ, on the other hand, contains over 30,000 high-resolution ($1024 \times 1024$) face images from more than 1,000 different celebrities. For our evaluation, we utilize a subset~\cite{celebAsub} of CelebA-HQ with 307 subjects that have been properly categorized.

{\bf Data Preprocessing.} Due to the extensive comparative content, we opt to use 10 subjects from this subset as our experimental subjects, ensuring diversity in terms of gender, ethnicity, and age. For each subject, four photos are selected and processed into $512 \times 512$ resolution.


{\bf Comparative attackers.} We compare CAAT with other attackers that are also applied in DMs, including Anti-DreamBooth (aspl) and Mist, which aims to protect users' portrait rights. In particular, to ensure fairness, the same four images are used as both the training and attacked images for Anti-DreamBooth (aspl). CAAT is compared with these two existing methods to evaluate their strengths and weaknesses.

{\bf Attacked model selection.} For customized fine-tuned models based on LDMs, four practical and popular ones, including Text Inversion, DreamBooth, Custom Diffusion, and SVDiff, are selected as the target models to achieve the adversarial examples through the attackers. This selection exhibits diversity and state-of-the-art performance. Successfully performing attacking on them can demonstrate that our CAAT can be effective on all LDMs-based fine-tuned models. Additionally, we compare the effectiveness of the attack on the variants of stable diffusion.

{\bf Evaluation metrics.} The quality of the generated face images is evaluated using the face detection and recognition model provided by InsightFace~\cite{insightface}. All the generated images are subjected to face detection to obtain the success rate of face detection called {\bf F}ace Detection Success {\bf R}ate {\bf (FR)}. For the generated images with detected faces, the average face similarity with four clean images is calculated, which is called {\bf F}ace {\bf S}imilarity {\bf (FS)}. FS takes values in the range of $[0,1]$, indicating the quality of generated images. Higher FS values signify lower face similarity and better attack effect. Moreover, {\bf I}mage{\bf R}eward {\bf (IR)}~\cite{xu2023imagereward} is employed  to compute the T2I generated image quality. It requires prompts corresponding to generated images, and evaluates the quality of images generated based on the prompts. Lower IR values signify lower image quality and better attack effect. Finally, we use {\bf F}réchet {\bf I}nception {\bf D}istance {\bf (FID)}~\cite{heusel2017gans} to measure the similarity between generated images and clean images. Higher FID indicates that the generated images are farther from clean images, signifying better attack effect.

{\bf Training setup.} We exclusively train CAAT on the $W_{K}$ and $W_{V}$ of cross-attention layers, using a batch size of 1 and a learning rate of $1\times10^{-5}$ for 250 training steps. Mixed precision with bf16 is employed. The attack prompt provided is ``a photo of a person". By default, we use the latest Stable Diffusion (v2.1) as the pretrained generator and set $\alpha$ to $5 \times 10^{-3}$ for CAAT, along with $\eta = 0.1$. Training CAAT with 500 steps on an NVIDIA RTX3090 takes approximately 2 minutes. We also summarize the hyperparameters for other attackers in \cref{tab:attacks} and DMs in \cref{tab:generate}, all of which are set to default values.


\subsection{T2I generation} 

First, the clean images are input into CAAT, Anti-DreamBooth, and Mist to obtain perturbed images (adversarial examples). Next, both the perturbed images and clean images undergo customized fine-tuning with Custom Diffusion, DreamBooth, SVDiff, and Textual Inversion. After the fine-tuning process, we generate 16 images with the prompt ``a photo of a person" using Stable Diffusion (v2.1). The experimental results are presented in \cref{tab:main_tab}, while visual representations of some results are shown in \cref{fig:compare}. As observed, CAAT successfully attacks all the models, yielding the best results for DreamBooth, Textual Inversion, and Custom Diffusion, and the second-best result for SVDiff. Although CAAT may not achieve optimal results across all evaluation metrics, the obtained values are already very low and visually imperceptible. Furthermore, both Anti-DreamBooth and Mist exhibit poor attack results on Custom Diffusion, underscoring CAAT's superior generalization capability. Additionally, while Mist achieves decent results, its added perturbation is more visually discernible, as evident in \cref{fig:perturb}. Moreover, we conducted additional experiments in Supplementary Material \hyperref[sec:More_Task]{B} with different prompts and subjects.

\subsection{Computational overhead}

\begin{figure}[]
          \centering
           \includegraphics[width=0.6\linewidth]{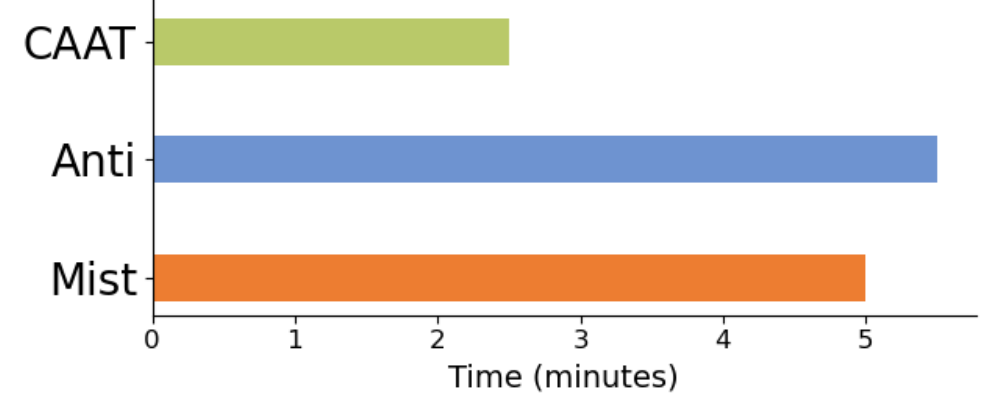}
        
            \caption{The training time of the attackers under the default settings of CAAT, Mist, and Anti-DreamBooth (Anti).}
           \label{fig:time}
        \end{figure}

We conducted analysis on computational overhead. The experiments were carried out by comparing CAAT, Mist, and Anti-DreamBooth under the same training setting. \Cref{fig:time} demonstrates our outstanding performance in terms of time efficiency. Training time of our method CAAT is about 2 minutes and 30 seconds on an NVIDIA RTX3090, compared to about 5 minutes and 30 seconds for Anti-DreamBooth and about 5 minutes for Mist on same GPU. CAAT is approximately twice faster than the other two. More importantly, CAAT does not require prior class images, but Anti-DreamBooth requires 200 images by default, which indicates that CAAT saves more cost.

\subsection{Ablation study}

We conduct ablation study to analyze the effect of CAAT. The experiments are carried by varying perturbation budgets, T2I diffusion models, and the quantity of perturbed images. 

{\bf Perturbation budgets.} We study the impact of different perturbation strengths when applying CAAT on the quality of T2I images. In our previous experiments, we set $\eta$ = 0.10, and now explore the effects of $\eta$ = 0.05 and $\eta$ = 0.15. The results are presented in \cref{tab:ablation_tab}. It can be observed that a larger $\eta$ leads to poorer T2I image quality, but excessively high $\eta$ settings introduce visually perceptible noise in the adversarial samples. \cref{fig:perturb} visually presents perturbed images generated by different attack methods. When using the default settings $\eta=0.10$, CAAT demonstrates superior attack effectiveness (as \cref{tab:main_tab}), with a level of noise similar to Anti-DreamBooth and less noise compared to Mist. However, when $\eta=0.15$, the perturbed images exhibit excessive noise.


{\bf T2I diffusion model variants.} It is essential to conduct the performance of the adversarial examples generated by CAAT on different versions of T2I diffusion models. In previous experiments, we apply CAAT on Stable Diffusion v2.1 for both the attack and T2I image generation. Additionally, we conducted experiments to assess the performance of CAAT's samples on Stable Diffusion v1.4 and v1.5, as shown in \cref{tab:SD}. CAAT demonstrates robust performance across different versions of Stable Diffusion, highlighting its strong generalization capabilities.


{\bf Quantity of perturbed images.} To simulate real-world scenarios where malicious attackers may obtain some clean images, we examine the impact of different proportions of perturbed images in the case of four input images. As indicated in \cref{tab:clean_perb}, the results demonstrate that more disturbed images lead to a more effective attack. CAAT consistently exhibits a robust attack effect, particularly with two or more perturbed images, whereas the impact is less pronounced with one or fewer perturbed images.

\subsection{Robustness of CAAT}
In real-world usage scenarios, images are easy to distortion, such as lossy compression or deformation. To verify if CAAT can handle complex real-world situations, we applied a variety of image perturbation methods in Supplementary Material \hyperref[sec:Robustness]{C} to demonstrate the robustness of CAAT.

\section{Conclusions}
\label{sec:conclusion}
We introduce a simple yet effective adversarial attack, CAAT, designed to protect users' portrait rights from infringement in the context of customized diffusion model fine-tuning. Users can employ CAAT to add imperceptible perturbations to images before publishing them, rendering malicious attackers unable to generate convincing fake images using these altered images. Our key idea is to introduce perturbations during the training of the cross-attention layers to disrupt the mapping between text and images. We conducte extensive experiments on DreamBooth, Textual Inversion, SVDiff, and Custom Diffusion, comparing the results to other attacks like Anti-DreamBooth and Mist. The result underscore the notable effectiveness and superior generalization capabilities of CAAT. Its ability to pre-process images for social media posts provides users with a powerful tool to fortify their portrait rights and protect against unauthorized image manipulations.

\section*{Acknowledgements}
This research was supported by the National Natural Science Foundation of China (No.62376023) and the Fundamental Research Funds for the Central Universities (Grant No.2022RC07X).

{
    \small
    \bibliographystyle{ieeenat_fullname}
    \bibliography{main}
}

\clearpage
\appendix

\renewcommand\thefigure{\Alph{section}\arabic{figure}}    
\renewcommand\thetable{\Alph{section}\arabic{table}}

\setcounter{page}{1}
\maketitlesupplementary

\section{Hyperparameters}
\label{sec:Hyperparameters}
\setcounter{table}{0}
\setcounter{figure}{0}
        \begin{table}[H]
          \centering
        \caption{Hyperparameters for different attackers.The parameters for Anti-DreamBooth (aspl) and Mist are set to their default configurations.  Anti. denotes Anti-DreamBooth. }
          \begin{tabular}{@{}cc@{}cc}
            \toprule
            parameters& CAAT & Anti&Mist\\
            \midrule
            train steps& 250& 50&100\\
            learning rate& $1 \times 10^{-5}$  \quad\quad & $5 \times 10^{-7}$&-\\
         $\alpha$& $5 
        \times 10^{-3}$\quad\quad & $5 
        \times 10^{-3}$&$2/255$\\
            $\eta$& 0.1& 0.05&$32/255$\\
            \bottomrule
          \end{tabular}
          \label{tab:attacks}
        \end{table}

        \begin{table}[H]
          \centering
          \caption{Hyperparameters of diffusion models, which follow their default configurations. CD, BD and TI denote Custom Diffusion, DreamBooth, and Textual Inversion, respectively.}
          \resizebox*{0.95\linewidth}{!}{
          \begin{tabular}{@{}cc@{}ccc}
            \toprule
            parameters& CD& DB&SVDiff& TI\\
            \midrule
            train steps& 250& 1000&500& 1500\\
            learning rate& $1 \times 10^{-5}$ \quad & $5 \times 10^{-7}$&$1 \times 10^{-3}$& $5 \times 10^{-4}$\\
         batchsize& 2& 1&1& 1\\
         \bottomrule
          \end{tabular}
          }
          \label{tab:generate}
        \end{table}

\section{More Task}
\label{sec:More_Task}
\setcounter{table}{0}
\setcounter{figure}{0}

\begin{figure*}
  \centering
  \includegraphics[width=1.0\linewidth]{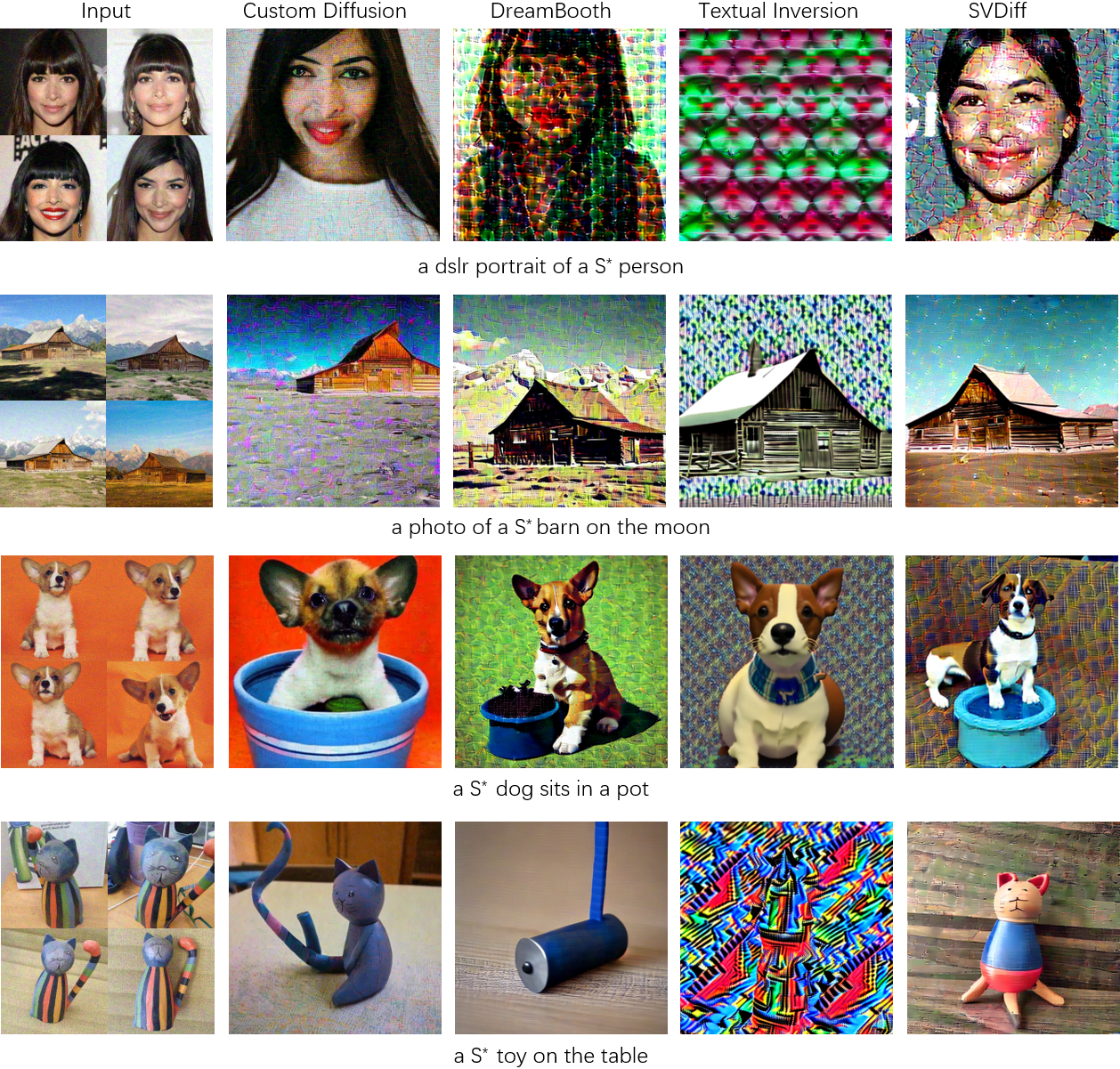}
   \caption{The images generated by different T2I diffusion models with different prompts and tasks. $S^*$ denotes special token of different T2I models.}
   \label{fig:rebbutal_1}
\end{figure*}

We studied the effects of different prompts on different subjects (e.g., barn, dogs, and toy). The results in \cref{fig:rebbutal_1} show no influence and that CAAT can consistently degrade the quality of generated images (first two lines have huge noise) and disrupt subject learning ability (the subjects of the last two lines are inconsistent).

\section{Robustness}
\label{sec:Robustness}
\setcounter{table}{0}
\setcounter{figure}{0}
We conducted more experiments with four image perturbation methods in \cref{tab:rebbutal} to demonstrate the robustness of CAAT. We used: \begin{itemize}
\item \textit{Random noise} has a scale of 0.05. 
\item \textit{Quantization} involves reducing an 8-bit image to a 6-bit image. 
\item \textit{Gaussian blur} uses a kernel size of 3x3 with $\sigma$ set to 0.05. 
\item \textit{JPEG} image processing is implemented using the OpenCV2 library.
\end{itemize}

\begin{table}[H]
\centering
\caption{Robustness assessment by different image perturbation methods. {\bf Bold} is the best score.}
\resizebox*{1.0\linewidth}{!}{
\begin{tabular}{c|cccccccc}
\hline
\multirow{3}{*}{Method} & \multicolumn{8}{c}{T2I generation models}                                                                                                         \\ \cline{2-9} 
                     & \multicolumn{4}{c|}{Custom Diffusion}                                             & \multicolumn{4}{c}{DreamBooth}                                \\
                     & FS$\downarrow$           & FC$\downarrow$             & IR$\downarrow$              & \multicolumn{1}{c|}{FID$\uparrow$}          & FS$\downarrow$             & FC$\downarrow$             & IR$\downarrow$              & FID$\uparrow$          \\ \hline
clean                & 1.0          & 0.52          & 0.47           & \multicolumn{1}{c|}{195}          & 0.98          & 0.52          & 0.53           & 179          \\ \hline
CAAT                 & 1.0 & 0.42          & \textbf{-0.36} & \multicolumn{1}{c|}{\textbf{250}} & \textbf{0.64} & \textbf{0.32} & \textbf{-0.14} & \textbf{371} \\ \hline
random noise         & 1.0          & 0.42          & -0.10          & \multicolumn{1}{c|}{222}          & 0.97          & 0.42          & 0.31           & 270          \\ \hline
quantization         & 1.0          & 0.44          & -0.15          & \multicolumn{1}{c|}{202}          & 0.99          & 0.45          & 0.42           & 201          \\ \hline
JPEG                 & 1.0          & \textbf{0.40} & -0.20          & \multicolumn{1}{c|}{218}          & 0.80          & 0.36          & 0.10           & 328          \\ \hline
Gaussian blur        & 1.0          & \textbf{0.40}          & -0.25          & \multicolumn{1}{c|}{229}          & 0.75          & 0.33          & -0.03          & 347          \\ \hline
\end{tabular}
}
\label{tab:rebbutal}
\end{table}

\section{Separation \vs simultaneous}
\label{sec:separated_optimization}
\setcounter{table}{0}
\setcounter{figure}{0}
When updating parameters and adding noise, we considered doing both simultaneously (See \cref{subsec:CAAT}) versus separately, aiming to find a superior method. For the latter, We alternated between 10 steps of model parameter updates and 10 steps of PGD , each for 250 steps (same as CAAT), with the results shown in \cref{tab:separated_optimization}. The experimental results indicate that both optimization methods achieved sufficiently good results, making it difficult to compare them. Moreover, for N-step training, simultaneous optimization requires N backward steps since we can reuse the gradients for attacking, while separation requires 2N. The goal of CAAT is to be lightweight and fast, introducing extra overhead is contrary to our philosophy. Therefore, we carried out the optimizations simultaneously.

\begin{table}[H]
\centering
\caption{Comparison of simultaneous optimization and separation optimization. \textit{Separated} involves alternating model optimization and adding noise.}
\resizebox*{1.0\linewidth}{!}{
\begin{tabular}{c|cccccccc}
\hline
\multirow{3}{*}{Method} & \multicolumn{8}{c}{T2I generation models}                                                                                                         \\ \cline{2-9} 
                     & \multicolumn{4}{c|}{Custom Diffusion}                                             & \multicolumn{4}{c}{DreamBooth}                                \\
                     & FS$\downarrow$           & FC$\downarrow$             & IR$\downarrow$              & \multicolumn{1}{c|}{FID$\uparrow$}          & FS$\downarrow$             & FC$\downarrow$             & IR$\downarrow$              & FID$\uparrow$          \\ \hline
clean                & 1.0          & 0.52          & 0.47           & \multicolumn{1}{c|}{195}          & 0.98          & 0.52          & 0.53           & 179          \\ \hline
CAAT                 & 1.0 & 0.42          & \textbf{-0.36} & \multicolumn{1}{c|}{\textbf{250}} & \textbf{0.64} & \textbf{0.32} & \textbf{-0.14} & \textbf{371} \\ \hline
Separated   & {\bf 0.99}            & {\bf 0.39}             & -0.25              & \multicolumn{1}{c|}{238}            & 0.72             & {\bf 0.32}             & -0.10              & 330            \\ \hline
\end{tabular}
}
\label{tab:separated_optimization}
\end{table}


\end{document}